\documentclass[sigconf]{acmart}
\usepackage{subfig}
\usepackage{multirow}
\usepackage{graphicx}
\usepackage{enumitem}

\AtBeginDocument{%
  }

\copyrightyear{2023}
\acmYear{2023}
\setcopyright{acmlicensed}
\acmConference[CIKM '23] {Proceedings of the 32nd ACM International Conference on Information and Knowledge Management}{October 21--25, 2023}{Birmingham, United Kingdom.}
\acmBooktitle{Proceedings of the 32nd ACM International Conference on Information and Knowledge Management (CIKM '23), October 21--25, 2023, Birmingham, United Kingdom}
\acmPrice{15.00}
\acmISBN{979-8-4007-0124-5/23/10}
\acmDOI{10.1145/3583780.3615280}

\usepackage{wrapfig}
\usepackage{booktabs}
\usepackage{xcolor}
\usepackage{enumerate}
\usepackage{url}            
\usepackage{booktabs}       
\usepackage{amsfonts}       
\usepackage{nicefrac}       
\usepackage{microtype}      
\usepackage{makecell, cellspace}         
\usepackage{color, colortbl}
\usepackage[utf8]{inputenc}
\usepackage{graphicx}
\usepackage{amsmath}
\usepackage{bbm}
\usepackage{amsthm}
\usepackage{algpseudocode}
\usepackage{multirow}
\usepackage{adjustbox}
\usepackage{enumitem}
\usepackage{bm}
\usepackage{mathtools}
\usepackage{mwe}
\usepackage{tabularx}
\usepackage{wrapfig}
\usepackage{capt-of}

\makeatletter
\newcommand*\bigcdot{\mathpalette\bigcdot@{.6}}
\newcommand*\bigcdot@[2]{\mathbin{\vcenter{\hbox{\scalebox{#2}{$\M\@th#1\bullet$}}}}}
\makeatother

\graphicspath{ {fig/} }

\newcommand{\proposed}{\textsf{S-Mixup}}

\begin{document}

\title{S-Mixup: Structural Mixup for Graph Neural Networks}

\author{Junghurn Kim}
\authornote{Both authors contributed equally to this research.}
\affiliation{%
  \institution{KAIST}
  \city{Daejeon}
  \country{Republic of korea}}
\email{jnghrn.j.kim@kaist.ac.kr}
  
\author{Sukwon Yun}
\authornotemark[1]
\affiliation{\institution{KAIST}
  \city{Daejeon}
  \country{Republic of korea}}
\email{swyun@kaist.ac.kr}
  
\author{Chanyoung Park}
\authornote{Corresponding author.}
\affiliation{\institution{KAIST}
  \city{Daejeon}
  \country{Republic of korea}}
\email{cy.park@kaist.ac.kr}

\begin{abstract}
\looseness=-1
Existing studies for applying the mixup technique on graphs mainly focus on graph classification tasks, while the research in node classification is still under-explored. In this paper, we propose a novel mixup augmentation for node classification called Structural Mixup (\proposed). The core idea is to take into account the structural information while mixing nodes. Specifically,~\proposed~obtains pseudo-labels for unlabeled nodes in a graph along with their prediction confidence via a Graph Neural Network (GNN) classifier. These serve as the criteria for the composition of the mixup pool for both inter and intra-class mixups. Furthermore, we utilize the edge gradient obtained from the GNN training and propose a gradient-based edge selection strategy for selecting edges to be attached to the nodes generated by the mixup. Through extensive experiments on real-world benchmark datasets, we demonstrate the effectiveness of~\proposed~evaluated on the node classification task. We observe that~\proposed~enhances the robustness and generalization performance of GNNs, especially in heterophilous situations. The source code of~\proposed~can be found at \url{https://github.com/SukwonYun/S-Mixup}.
\end{abstract}

\vspace{-1.5ex}


\vspace{-2ex}


\maketitle


\section{Introduction}

\begin{figure}[t]
    \includegraphics[width=0.9\columnwidth]{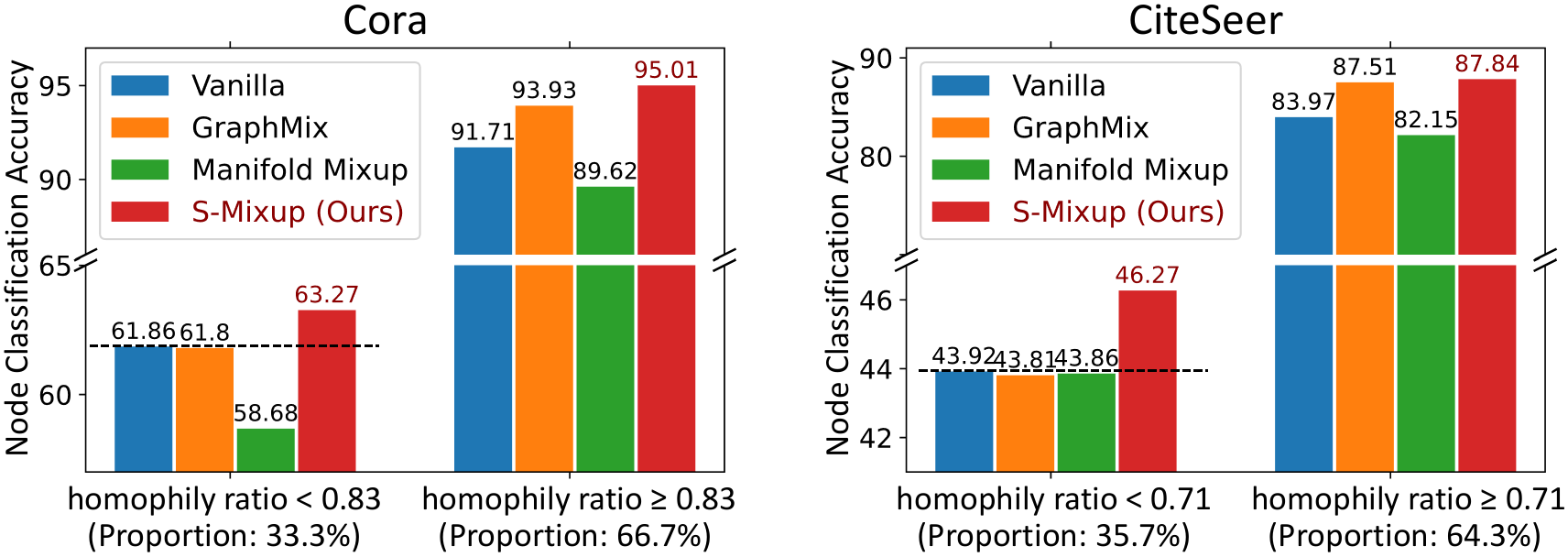}
    \vspace{-2.5ex}
    \caption{Comparison of performance between nodes with relatively low and high homophily ratios in the Cora and CiteSeer datasets. Criterion: average homophily ratio.} \label{fig:intro_fig1}
    \vspace{-4ex}
\end{figure}

Despite the notable successes of Mixup-based data augmentation in computer vision~\cite{ImageMixup,Cutmix,PuzzleMix}, the application of mixup in the graph domain remains under-explored due to the challenges that naturally arise within graph data, i.e., \textit{existence of structure information}. As the recent success of Graph Neural Networks (GNNs)~\cite{chebnet, gcn, gat, graphsage, gin} stems from their capacity to utilize both feature and structural information, it is essential to integrate structural information alongside feature information for mixup in graphs. For this reason, a naive application of existing methods in computer vision, which primarily focus on feature information, would inevitably yield suboptimal results. Several recent studies have implemented mixup techniques on graphs, including GraphTransplant~\cite{GraphTransplant}, G-Mixup~\cite{GMixup}, IfMixup~\cite{IfMixup}, and Manifold Mixup~\cite{ManifoldNodeMixup}. These studies commonly propose to interpolate a subset of nodes in the representation space, and have demonstrated effectiveness mainly in \textit{graph classification} tasks.




However, only a handful of studies address mixup on graph from a single node perspective, i.e., \textit{node classification}, which is a representative downstream task in the graph domain.
Compared to mixup models in graph classification, where a graph representation is simply obtained by pooling over all nodes, executing mixup in node classification poses an inherent challenge due to its direct influence on neighboring nodes, i.e., the impact of closely-located structural information becomes more crucial. Although GraphMix~\cite{GraphMix} and Manifold Mixup~\cite{ManifoldNodeMixup} address the node classification task by proposing Fully-Connected Networks for node mixing and a two-branch design for target and local neighbors, respectively, they still insufficiently utilize the structural information within the mixed nodes. This could inadvertently introduce and convey irrelevant information during the message-passing process.

To delve deeper into structural information, one characteristic that intuitively represents such property from a node's perspective is the \textit{homophily ratio}, i.e., the ratio of edges that share the same class as the target node to the total number of neighboring edges. In an ideal mixup process that incorporates this structural information, unbiased performance gains should occur between homophilous and heterophilous scenarios, enhancing results across both contexts. To verify, we evaluate the node classification accuracy of node-perspective mixup models, GraphMix~\cite{GraphMix} and Manifold Mixup~\cite{ManifoldNodeMixup}, against a vanilla Graph Convolutional Network (GCN). We perform this evaluation with respect to nodes exhibiting relatively low homophily ratio (i.e., $< 0.83$), as well as those with high homophily ratio (i.e., $\geq 0.83$) (see Figure~\ref{fig:intro_fig1}). Interestingly, our analysis in Cora and CiteSeer citation networks reveals GraphMix's performance gain stems from accurately classifying high homophily nodes. However, we observe a performance drop for nodes with a relatively low homophily ratio compared to the vanilla GCN baseline. {This suggests that newly generated nodes through mixup do not significantly contribute to smoothing decision boundaries and achieving generalizability.} Moreover, given nodes with low homophily ratio comprise around 1/3 of all nodes (33.3\% in Cora and 35.7\% in Citeseer), their relatively low performance should not be ignored. This becomes critical in real-world scenarios, where disassortative nodes inevitably emerge within assortative graphs, especially when such nodes are linked to malicious activities like fraud or bots \cite{bot,lte4g} negatively affecting neighbors. Hence, a balanced approach that prevents bias towards high homophily nodes, considering structural information, is vital for optimal performance.


\looseness=-1
Furthermore, existing methods such as GraphMix and Manifold Mixup lack \textit{the ability to manage newly generated nodes} in terms of their structural information. These methods primarily focus on obtaining representations of mixed nodes without considering their local neighborhood context. As a result, they resort to replacing existing nodes with newly generated ones instead of adding them to the graph and connecting them to existing nodes. This approach loses the opportunity to utilize the original features of the given nodes and leverage message-passing through the newly generated edges.

In this regard, with the goal of properly utilizing structural information, we propose a novel mixup method for node classification, called Structural Mixup (\proposed), which equips the ability to connect the relevant edges to the newly generated nodes that seamlessly align with the current graph. More precisely, we first pass the original graph through the GNN classifier to obtain two key components that enable the mixup process while incorporating structural information: pseudo-labels with prediction confidence and edge gradients. We utilize pseudo-labels with prediction confidence to expand the candidate pool for node mixup. Specifically, nodes with high and low prediction confidence are selected for the intra-class mixup, while nodes with medium confidence are used for the inter-class mixup. We then utilize the edge gradients to identify edges with high gradient values. The newly generated nodes are connected to the existing nodes and are passed through the GNN classifier. Extensive experiments illustrate that~\proposed~outperforms existing mixup-based GNNs in node classification task.





\vspace{-2.5ex}

\section{Methodology}
\noindent{\textbf{Notations.}} Given a graph $\mathcal{G}=(\mathcal{V},\mathcal{E},\mathbf{X})$, let $\mathcal{V}=\{v_1,...,v_N\}$ denote the set of nodes, $\mathcal{E} \subseteq \mathcal{V} \times \mathcal{V}$ denote the set of edges, and $\mathbf{X} \in \mathbb{R}^{N \times F}$ denote the node feature matrix. We use $\mathcal{C}$ to denote the set of classes of nodes in $\mathcal{G}$. We define $\mathbf{A} \in \mathbb{R}^{N \times N}$ as the adjacency matrix where $\mathbf{A}_{ij}=1$ iff $(v_i, v_j) \in \mathcal{E}$ and $\mathbf{A}_{ij}=0$ otherwise. Our main goal is to enhance performance in a node classification task.

\begin{figure}[t]
    \centering
    \includegraphics[width=0.9\columnwidth]{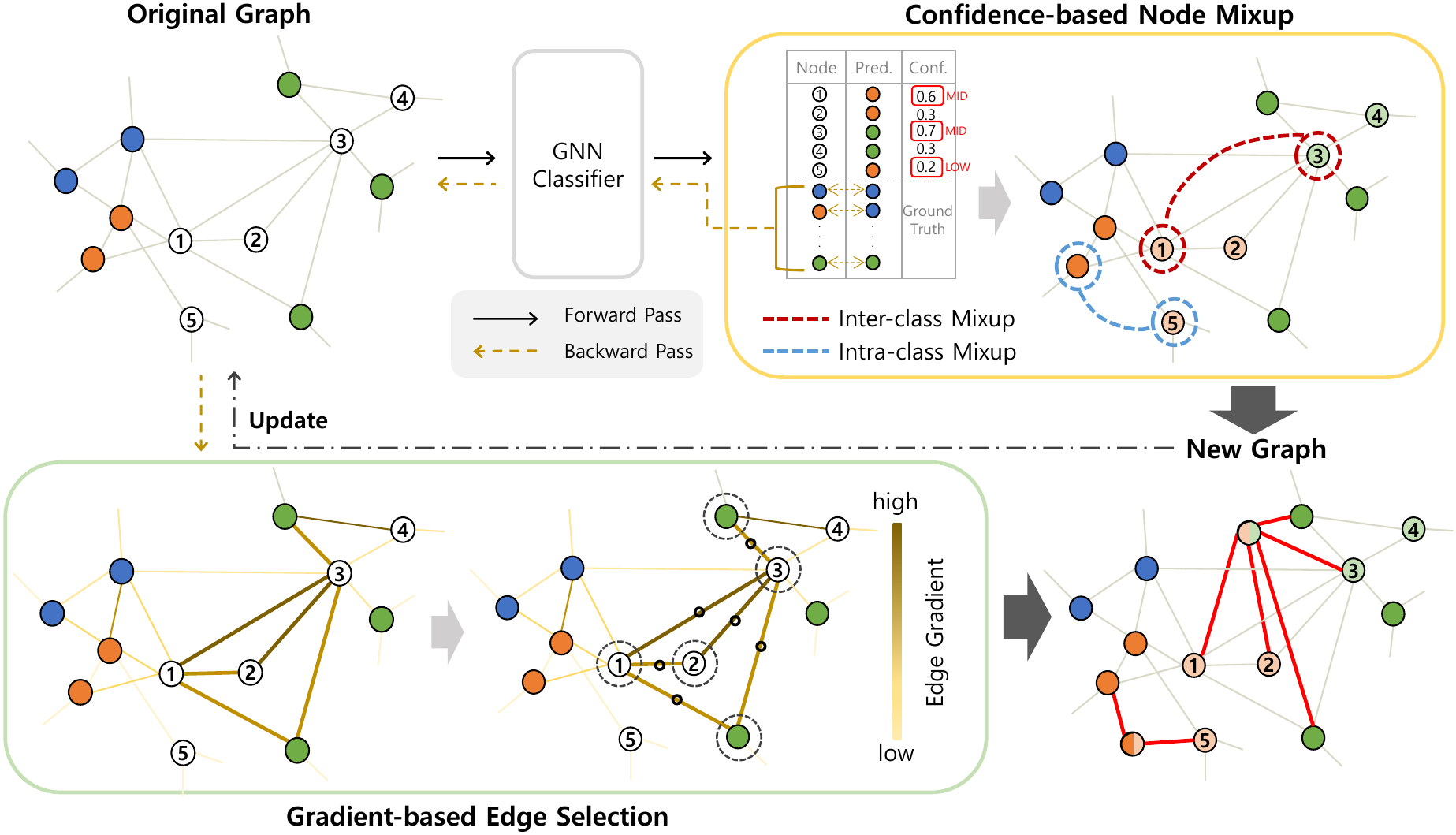}
    \vspace{-2.5ex}
    \caption{Overall architecture of~\proposed.}
    \vspace{-3.7ex}
    \label{fig:main_figure}
\end{figure}

\noindent{\textbf{Our approach.}} We propose a novel mixup method, Structural Mixup (\proposed), that incorporates structural information within newly generated nodes. We first revisit vanilla GNN to obtain two key outputs, pseudo-labels with prediction confidence and edge gradient, in Section 3.1. Then, we describe how we expand the mixup pool through the prediction confidence and conduct inter-class and intra-class mixups in Section 3.2. With edge gradients obtained via a vanilla GNN, we then demonstrate how we select edges of the newly generated nodes in Section 3.3. Finally, we propose an overall training process of~\proposed~in Section 3.4. The overall architecture of~\proposed~is illustrated in Figure~\ref{fig:main_figure}.

\vspace{-2ex}

\subsection{Revisiting Graph Neural Networks}
Before we perform node mixup and edge selection, we first revisit the vanilla GNN to obtain key outputs that will play a significant role in the following sections. Specifically, we pass a graph through a conventional two-layer Graph Convolutional Network \cite{gcn} as: $\hat{\mathbf{Y}} = \text{softmax}(\hat{\mathbf{A}} \sigma(\hat{\mathbf{A}}\mathbf{X}\mathbf{W}^{1})\mathbf{W}^{2})$,
where $\hat{\mathbf{Y}} \in \mathbb{R}^{N\times |\mathcal{C}|}$ denotes prediction probability, 
$\hat{\mathbf{A}} = \tilde{\mathbf{D}}^{-1/2}\tilde{\mathbf{A}}\tilde{\mathbf{D}}^{-1/2}$ denotes a transition matrix where $\tilde{\mathbf{D}}$ and $\tilde{\mathbf{A}}$ indicates the self-loop included diagonal degree matrix and adjacency matrix, respectively. $\sigma$ is the ReLU activation function, and $\mathbf{W}^{1}\in \mathbb{R}^{F\times D}$ and $ \mathbf{W}^{2}\in \mathbb{R}^{D\times |\mathcal{C}|}$ denote trainable weight matrices that encode features into hidden embedding space, $D$ and hidden embedding space to class space $|\mathcal{C}|$, respectively. Here, we define one-hot transformed \textit{pseudo-labels} of each node as $\tilde{\mathbf{Y}} = \text{one-hot}(\text{argmax}(\hat{\mathbf{Y}})) \in \mathbb{R}^{N\times |\mathcal{C}|}$, and \textit{prediction confidence} of each node as $\mathbf{y}_\text{conf} = \text{max}(\hat{\mathbf{Y}}) \in \mathbb{R}^{N}$, where $0 \leq \mathbf{y}_\text{conf} \leq 1$ due to the softmax operation. This will later serve as a criterion for node mixup. Then the node classification loss is calculated as:
$\mathcal{L}_{\text{ce}} = -\sum_{v \in \mathcal{V}_{tr}} \sum_{c \in \mathcal{C}} \mathbf{y}_{v}[c]\log(\hat{\mathbf{y}}_{v}[c])$,
where $\mathcal{V}_{tr}$ denotes the set of training nodes and $\mathbf{y}_{v}[c]$ is either 1 or 0 depending on the value in the $c$-th index of class one-hot vector of training node $v$. During backpropagation using gradient descent within this loss, we can naturally obtain gradients with respect to an adjacency matrix, $\tilde{\mathbf{A}}$ by applying the chain rule as follows:
\begin{equation}
\small
        \frac{\partial \mathcal{L}_{\text{ce}}}{\partial \tilde{\mathbf{A}}} = \frac{\partial \mathcal{L}_{\text{ce}}}{\partial \hat{\mathbf{Y}}} \frac{\partial \hat{\mathbf{Y}}}{\partial \hat{\mathbf{A}}} \frac{\partial \hat{\mathbf{A}}}{\partial \tilde{\mathbf{A}}} 
     = -(\hat{\mathbf{Y}}^{-1})\cdot\{(\hat{\mathbf{A}}\mathbf{X}\mathbf{W}^{1}\mathbf{W}^{2})^{\top}\mathbf{I} + (\mathbf{X}\mathbf{W}^{1}\mathbf{W}^{2})^{\top}\hat{\mathbf{A}}\}\cdot(\tilde{\mathbf{D}}^{-1})
\end{equation}
It is important to note that while the adjacency matrix does not possess trainable edge weights and thus maintains a static state, its gradient does change with each epoch due to the modifications in the trainable weight parameters $\mathbf{W}^{1}$ and $\mathbf{W}^{2}$. Here, we define the \textit{edge gradient} for each edge as $e_{ij} = \left|\frac{\partial L}{\partial \tilde{\mathbf{A}}}\right|_{2}, \forall i,j \leq N$, which will later serve as a criterion for edge selection.

\subsection{Confidence-based Node Mixup}
Given the \textit{pseudo-labels} $\tilde{\mathbf{Y}}$ with \textit{prediction confidence} $\mathbf{y}_\text{conf}$, we now perform mixup while considering the class information as well as its uncertainty, i.e., confidence in terms of Inter-class Mixup and Intra-class Mixup. In essence, during the Inter-class Mixup, we aim to use nodes from different classes that have medium-level confidence. This approach helps to smooth the decision boundary between classes. On the other hand, during the Intra-class Mixup, we aim to use nodes from the same class with high and low confidence levels. This approach enhances generalizability within a class.


\smallskip
\noindent{\textbf{Inter-class Mixup. }} The key success of mixup lies in its ability to create smoother decision boundaries, which are well-established factors contributing to a model's generalizability~\cite{smooth1, smooth2}. Thus, to achieve such smooth decision boundaries among different classes, we utilize nodes with medium-level confidence. The intuition behind using medium-level confidence nodes is that nodes with high confidence tend to be far from the decision boundary and too discriminative (i.e., easy to classify), whereas nodes with low confidence may be too close to the decision boundary (i.e., hard to classify). Hence, we choose nodes with medium-level confidence for obtaining smooth boundaries between classes. Formally, we consider the middle 2$r\%$ among $\mathbf{y}_\text{conf}$ in each class as the inter-class mixup pool, where $r$ is a hyperparameter. The generation of new nodes can then be achieved by randomly sampling from the inter-class mixup pool in each class, as follows:
\begin{equation}
\label{eq:inter_class_x}
\small
  \mathbf{{x}}_\text{new}^\text{Inter} = \lambda \mathbf{x}_{i} + (1 - \lambda) \mathbf{x}_{j}, \,
  \mathbf{{y}}_\text{new}^\text{Inter} = \lambda \mathbf{\tilde{y}}_{i} + (1 - \lambda) \mathbf{\tilde{y}}_{j}, \,
  \forall i,j \leq N \; \textit{s.t.  }   \tilde{\mathbf{y}}_i \neq \tilde{\mathbf{y}}_j
\end{equation}

\noindent where $\mathbf{{x}}_\text{new}^\text{Inter} \in \mathbb{R}^{F}$, $\mathbf{{y}}_\text{new}^\text{Inter} \in \mathbb{R}^{|\mathcal{C}|}$ denote the feature, label of the newly generated node, respectively, and $\lambda \in [0,1]$ is the mixing ratio, drawn from a Beta distribution with parameter $\alpha$ fixed as 1.0.


\smallskip
\noindent{\textbf{Intra-class Mixup. }} At the same time, considering nodes within the same class is also crucial, as it can significantly influence the decision boundaries of the classes. Thanks to the pseudo-labels we previously obtained, we can now conduct intra-class mixup to achieve robust characteristics for each class. Specifically, we aim to obtain robust features for each class by interpolating between nodes with high confidence and those with low confidence. Formally, we consider the upper $r\%$ among $\mathbf{y}_\text{conf}$ as the mixup pool for high-confidence nodes and the lower $r\%$ among $\mathbf{y}_\text{conf}$ as the mixup pool for low-confidence nodes. Then, through random sampling from the intra-class mixup pool, i.e., high-confidence and low-confidence mixup pools, we generate new nodes in each class as follows:
\begin{equation}
\small
\label{eq:intra_class_x}
  \mathbf{{x}}_\text{new}^\text{Intra} = \lambda \mathbf{x}_{i} + (1 - \lambda) \mathbf{x}_{j},  \,
  \mathbf{{y}}_\text{new}^\text{Intra} = \mathbf{\tilde{y}}_{i} = \mathbf{\tilde{y}}_{j}, \,
  \forall i,j \leq N \; \textit{s.t.  }   \tilde{\mathbf{y}}_i = \tilde{\mathbf{y}}_j
\end{equation}
where $\mathbf{{x}}_\text{new}^\text{Intra} \in \mathbb{R}^{F}$ denotes the feature of the newly generated node, and $\mathbf{{y}}_\text{new}^\text{Intra}$ shares the same one-hot class vector as nodes $i$ and $j$. 

\begin{figure}[!t]
    \includegraphics[width=0.9\columnwidth]{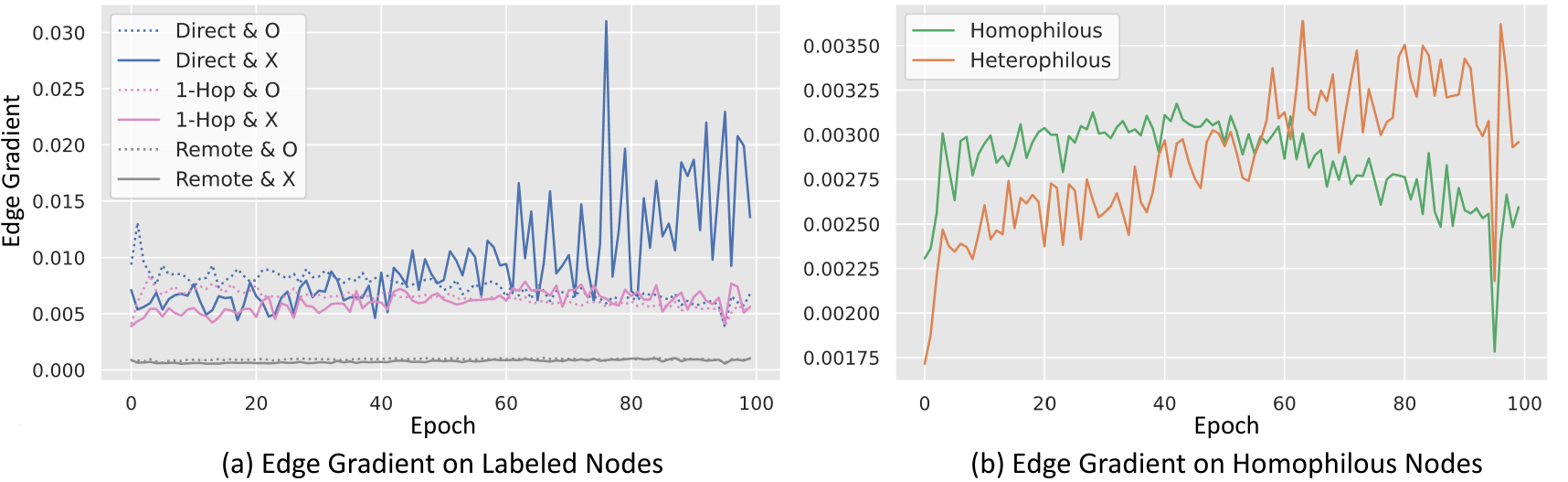}
    \vspace{-2.5ex}
    \caption{Edge gradient of vanilla GCN in Cora dataset. (a) `Direct \& O', `1-Hop \& X' represent a correctly predicted edge directly linked to labeled nodes and an incorrectly predicted edge linked to labeled nodes within 1-hop, respectively. (b) `Homophilous' denote edges connecting same labeled nodes.}
    \vspace{-4ex}
    \label{fig:gradient}
\end{figure}


\begin{table*}[t]
\begin{minipage}{0.68\linewidth}{
\centering
    \includegraphics[width=0.95\columnwidth]{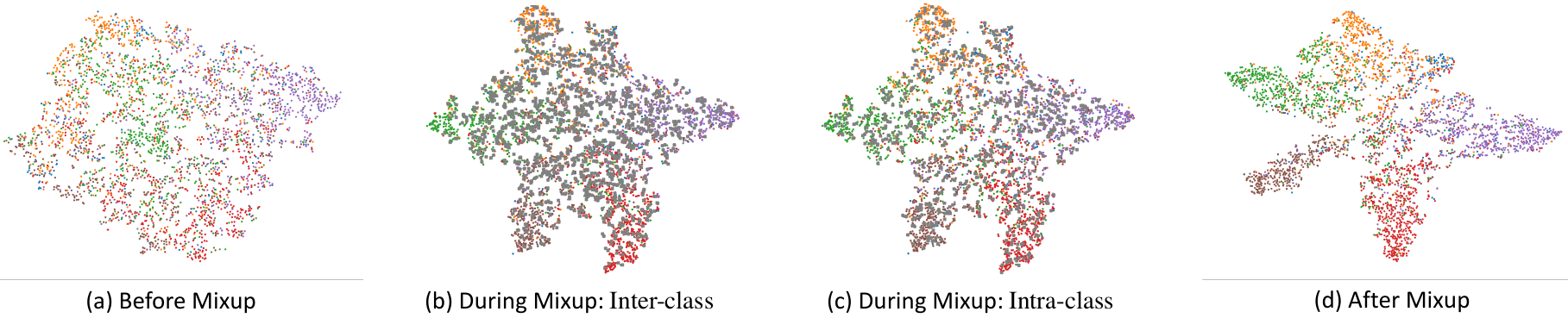}
    \vspace{-2ex}
    \captionof{figure}{t-SNE representation of nodes during training~\proposed~in CiteSeer dataset. \textcolor{gray}{Gray color: newly generated mixup nodes.}}
    \label{fig:tsne}
}\end{minipage}
\begin{minipage}{0.31\linewidth}{
\centering
    \includegraphics[width=0.95\columnwidth]{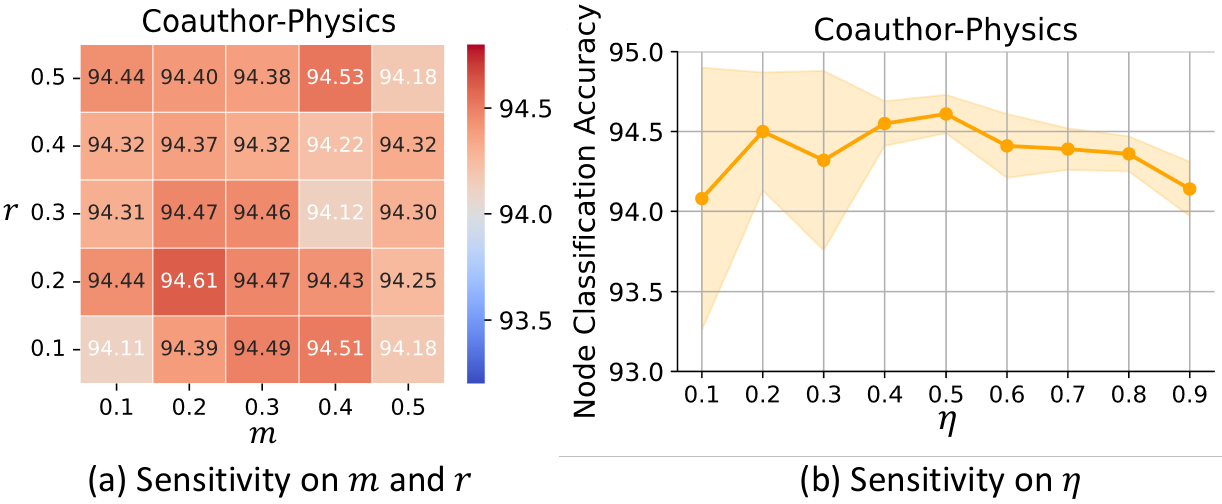}
    \vspace{-2.5ex}
    \captionof{figure}{Sensitivity on $r$, $m$, $\eta$.}
    \label{fig:sensitivity}
}\end{minipage}
\vspace{-2ex}
\end{table*}




\subsection{Gradient-based Edge Selection}
Recall that we are dealing with the graph-structured data where both nodes and edges play significant roles. In this regard, the subsequent challenge that naturally emerges is: \textit{how do we create connections between the newly generated nodes and the existing ones?} To address this question, we propose to leverage the edge gradient, $e_{ij}$, which we acquired during the vanilla GNN in Section 3.1. The rationale behind utilizing the edge gradient is twofold: \textbf{(1)} The newly generated nodes would be in favor of being connected to nodes that can convey supervisory signals to them. Intuitively, as the training loss is derived from labeled training samples, the gradient value within an edge that is either directly connected to or located within a 1-hop distance from the original training nodes would be larger than that of an edge that is neither connected nor in vicinity of the training nodes. This tendency is corroborated by Figure~\ref{fig:gradient} (a), where we report the edge gradient with respect to labeled nodes as the training epoch increases. In short, an edge with high gradient implies that it is close to the labeled nodes, and thus such an edge would be preferred for newly generated nodes in order to transmit supervised signals. \textbf{(2)} The existing nodes would prefer to be connected to newly generated nodes if these new connections can help alleviate their local and structural difficulties, particularly in heterophilous scenarios. More precisely, nodes that are surrounded by disassortative neighbors would encounter difficulties in being correctly classified, thus necessitating mitigation. 
This phenomenon is substantiated by Figure~\ref{fig:gradient} (b), which illustrates the edge gradient dynamics for both homophilous and heterophilous situations. We observe that with increasing epochs, the edge gradient value intensifies for difficult samples, i.e., those on heterophilous edges. Consequently, for existing nodes, selecting high-gradient edges emerges as a preferable strategy for existing nodes, as it provides opportunities to better manage such heterophilous situations. To sum up, both newly generated nodes and existing nodes would benefit from being connected to edges with high gradient values. 



Now, to create edges for the newly generated nodes $\mathbf{x}_{\text{new}}^{\text{Inter}}, \mathbf{x}_{\text{new}}^{\text{Intra}}$, we leverage the edge gradient obtained during the training of the vanilla GNN. In each epoch, we initially connect a generated node with the two nodes from which the node is generated. Subsequently, we expand the connections to encompass high-gradient edges that fall within the top $m\%$ of the total edge gradients with hyperparameter $m$ and are connected to either of the two nodes involved in the node mixup. {Formally, the adjacency for a newly generated node stemming from the source node $i$ and $j$ can be expressed as follows:}
\begin{equation}
\small
\label{eq:edge_connector}
\mathbf{A}_{\text{new},k} = 
\begin{cases}
    1, & \text{if} \;\; m\text{-th Percentile}(e_{\cdot\cdot}) \leq e_{kq}, \; \forall k \leq N \; \text{s.t. } q \in \{i,j\} \\
    0, & \text{otherwise}
\end{cases}
\end{equation}
where $e_{\cdot\cdot}$ denote the list of all edge gradient values and $e_{kq}$ denote edge gradient value within node $k$ and node $q$.

\subsection{Model Training}
To sum up, the overall training process of~\proposed~is expressed as: 
$\mathcal{L}_\text{final} = \mathcal{L}_\text{ce} + \eta\mathcal{L}_\text{mixup}^{\text{Intra}} + (1-\eta)\mathcal{L}_\text{mixup}^{\text{Inter}}$,
where $\mathcal{L}_\text{ce}$ denotes the cross entropy loss computed from the GNN classifier with the labeled training nodes, and $\mathcal{L}_\text{mixup}^{\text{Intra}}, \mathcal{L}_\text{mixup}^{\text{Inter}}$ represent the cross entropy loss computed from GNN classifier with the newly added nodes with their corresponding pseudo-labels in intra-class mixup and inter-class, respectively, with a loss balancing hyperparameter $\eta$. During training, since node mixup and edge connection are implemented based on the pseudo-labels and prediction confidence derived from the initial GNN, we prioritize the optimization of $\mathcal{L}_\text{ce}$ loss followed by optimizing $\eta\mathcal{L}_\text{mixup}^{\text{Intra}}+(1-\eta)\mathcal{L}_\text{mixup}^{\text{Inter}}$ loss in our implementation.



\vspace{-2ex}
\begin{table}[t]
\caption{Data statistics.}
\vspace{-3ex}
\resizebox{0.9\columnwidth}{!}{
\begin{tabular}{@{}l|ccccccc@{}}
\toprule
\multicolumn{1}{c|}{\textbf{Dataset}} & \textbf{\# Nodes} & \textbf{\# Edges} & \textbf{\# Features} & \textbf{\# Cls.} & \textbf{\# Train} & \textbf{\# Val.} & \textbf{\# Test} \\ \midrule
Cora                         & 2,708    & 5,429    & 1,433       & 7         & 140        & 500          & 1,000      \\
CiteSeer                     & 3,327    & 4,732    & 3,703       & 6         & 120        & 500          & 1,000      \\
PubMed                       & 19,717   & 44,338   & 500        & 3         & 60         & 500          & 1,000       \\
Co. CS                       & 18,333   & 81,894   & 6,805       & 15        & 300        & 450          & 17,583       \\
Co. Physics                  & 34,493   & 247,962  & 8,415       & 5         & 100        & 150          & 34,243      \\ \bottomrule
\end{tabular}
}

\vspace{-2ex}
\label{tab:data_statistics}
\end{table}

\section{Experiments}

\noindent{\textbf{Datasets. }} We evaluate~\proposed~on five benchmark citation datasets, namely Cora, CiteSeer, PubMed, Coauthor-CS, and Coauthor-Physics. For the split, we follow the data split setting of GraphMix~\cite{GraphMix}. The detailed statistics can be found in Table~\ref{tab:data_statistics}. 

\smallskip
\noindent{\textbf{Compared Methods. }} We compare~\proposed~with GCN~\cite{gcn}, node-perspective mixup models such as GraphMix~\cite{GraphMix} and Manifold Mixup~\cite{ManifoldNodeMixup}, as well as an oversampling-based model, GraphSMOTE~\cite{graphsmote}, and a feature saliency-based mixup model, GraphENS~\cite{graphens}.

\smallskip
\noindent{\textbf{Evaluation Protocol. }} All experiments are repeated ten times with randomly initialized parameters, and we present the mean accuracy and standard deviation conducted on RTX 3090 (24GB). Common hyperparameters include a learning rate chosen from $\{0.01, 0.05, 0.1\}$, hidden dimensions from $\{16, 64\}$, and a fixed dropout rate of 0.5. For each model, individual hyperparameters are tuned within a range as recommended by the respective authors.

\smallskip
\noindent{\textbf{Overall Performance. }} In Table~\ref{tab:overall_performance},~\proposed~outperforms other mixup-based models such as GraphMix and Manifold Mixup, as well as oversampling-based models like GraphSMOTE and GraphENS. The vanilla GCN model serves as the lower performance bound. Specifically, in CiteSeer dataset, our method achieves the highest performance, showing a significant improvement of 4.69\% over the vanilla GCN, which is a substantial improvements considering that 
the performance gain of GraphMix over the vanilla GCN is 2.35\%.

\smallskip
\noindent{\textbf{Ablation Studies. }} As~\proposed~employs a confidence-based node mixup for both inter-class and intra-class nodes with a gradient-based edge selection strategy, we observe several key outcomes from Table~\ref{tab:overall_performance}. \textbf{(1)} Incorporation of both inter-class and intra-class mixup is crucial to the efficacy of the node mixup technique. This integration aids not only in generating smoother decision boundaries but also in establishing a more generalized representation for each class. Figure~\ref{fig:tsne} provides a detailed view of the progression of node representations as epochs grow. We observe that generated nodes in both the inter-class and intra-class mixup are placed near the decision boundaries, which aligns with the objectives of smoothing decision boundaries and creating generalized representations. \textbf{(2)} It is essential to connect the newly generated nodes with the pre-existing nodes. Our findings show that even when the proposed method is used with a low edge gradient (\proposed~w/ low Edge), it surpasses the performance of the same method without any edge connection (\proposed~w/o Edge). 
\textbf{(3)} Our proposed method is superior when using high-gradient edges rather than low-gradient ones. High-gradient edges encapsulate a greater degree of supervised information, providing an opportunity to mitigate heterophilous situations, as mentioned in Section 3.3.

\vspace{-0.8ex}


\smallskip
\noindent{\textbf{Sensitivity Analysis. }} Figure~\ref{fig:sensitivity} (a) demonstrates the hyperparameter sensitivity with respect to the sampling hyperparameter $r\in\{0.1,0.2,0.3,0.4,0.5\}$, which is responsible for the inter- and intra-class mixup, and $m\in\{0.1,0.2,0.3,0.4,0.5\}$ that dictates high edge gradient selection. Interestingly, our proposed method exhibits robustness against significant variations in both hyperparameters. However, favoring a joint small range for $m$ and $r$ appears to be beneficial. This is likely because node generation via high-confidence values (i.e., small $r$) and node connections via high edge gradient values (i.e., small $m$) introduce less uncertainty compared to their low-confidence and low-gradient counterparts. In Figure~\ref{fig:sensitivity} (b), we also investigate the impact of the loss controlling parameter $\eta\in\{0.1,0.2,...,0.9\}$, keeping $m$ and $r$ fixed. We find that a balanced consideration of both inter- and intra-class mixup, represented by $\eta=0.5$, is advantageous for training~\proposed.

\begin{table}[t]
\caption{Overall performance with ablation studies.}
\vspace{-3ex}
\label{tab:overall_performance}
\resizebox{0.93\columnwidth}{!}{
\begin{tabular}{@{}l|ccccc@{}}
\toprule
\multicolumn{1}{c|}{\textbf{Models}} & \textbf{Cora} & \textbf{CiteSeer} & \textbf{PubMed} & \textbf{Co. CS} & \textbf{Co. Physics} \\ \midrule
Vanilla                                & 81.82\small{$\pm$0.43}    & 69.70\small{$\pm$0.76}        & 78.62\small{$\pm$0.38}      & 91.45\small{$\pm$0.31}      & 93.46\small{$\pm$0.21}           \\
GraphMix                               & 83.31\small{$\pm$0.60}    & 72.04\small{$\pm$0.52}        & 77.95\small{$\pm$1.09}      & 92.18\small{$\pm$0.11}      & 94.35\small{$\pm$0.08}           \\
Manifold Mixup                    & 79.52\small{$\pm$0.86}    & 68.99\small{$\pm$0.81}        & 76.63\small{$\pm$1.10}      & 90.30\small{$\pm$0.50}      & 92.49\small{$\pm$0.47}           \\
$\text{GraphSMOTE}_{T}$                           & 80.76\small{$\pm$0.79}    & 69.61\small{$\pm$0.93}        & 79.62\small{$\pm$1.09}      & 90.61\small{$\pm$0.35}      & OOM                  \\
$\text{GraphSMOTE}_{O}$                           & 80.76\small{$\pm$0.85}    & 69.02\small{$\pm$1.50}        & 79.58\small{$\pm$0.97}      & 90.52\small{$\pm$0.38}      & OOM                  \\
GraphENS                               & 81.95\small{$\pm$0.59}    & 72.73\small{$\pm$0.49}        & 78.78\small{$\pm$0.38}      & 91.26\small{$\pm$0.26}      & OOM                  \\ \midrule
\proposed~w/o Inter Node                 & 84.12\small{$\pm$0.94}    & 73.43\small{$\pm$2.46}        & 78.48\small{$\pm$0.98}      & 91.72\small{$\pm$0.40}      & 93.92\small{$\pm$0.20}           \\
\proposed~w/o Intra Node                 & 82.88\small{$\pm$2.61}    & 70.90\small{$\pm$2.39}        & 78.84\small{$\pm$1.01}      & 92.10\small{$\pm$0.28}      & 94.54\small{$\pm$0.28}           \\
\proposed~w/o Edge                       & 84.10\small{$\pm$0.75}    & 70.96\small{$\pm$3.91}        & 78.85\small{$\pm$1.35}      & 91.92\small{$\pm$1.14}      & 94.05\small{$\pm$0.49}           \\
\proposed~w/ low Edge                    & 84.15\small{$\pm$0.89}    & 72.59\small{$\pm$3.90}        & 78.90\small{$\pm$0.64}      & 92.09\small{$\pm$0.23}      & 94.24\small{$\pm$0.25}           \\ \midrule
\proposed~(Ours)                         & \textbf{84.78}\small{$\pm$0.42}    & \textbf{74.39}\small{$\pm$0.65}        & \textbf{79.70}\small{$\pm$0.39}      & \textbf{92.33}\small{$\pm$0.18}      & \textbf{94.61}\small{$\pm$0.12}           \\ \bottomrule
\end{tabular}
}
\vspace{-2ex}
\end{table}

\section{Conclusion}
In this paper, we present a novel mixup augmentation technique that thoroughly accounts for the structural information inherent in the graph domain. We identified that existing works, which do not fully incorporate structural information, fail to generalize well in heterophilous scenarios. In response to this,~\proposed~initially processes the vanilla GNN to acquire pseudo-labels along with their prediction confidence. Subsequently, we constitute the mixup pool for both inter-class and intra-class cases using nodes with medium levels of prediction confidence, and a combination of high and low confidence levels, respectively. Furthermore, to establish connections between newly generated nodes and existing ones, we employ the edge gradient-based edge selection. Through comprehensive experiments, we showed the superiority of~\proposed~in enhancing the robustness and generalization of node classification task.

\smallskip

\noindent\textbf{Acknowledgement}: 
No.2021R1C1C1009081 and No.2022-0-00157.

\bibliographystyle{ACM-Reference-Format}
\bibliography{CIKM_mixup}





\end{document}